\documentclass[10pt,twocolumn,letterpaper]{article}
\usepackage{geometry}
\usepackage{bbding}
\usepackage{times}
\usepackage{epsfig}
\usepackage{graphicx}
\usepackage{amsmath,amssymb}
\usepackage{booktabs}
\usepackage{multirow}
\usepackage{natbib}
\usepackage[table]{xcolor}
\usepackage[ruled,vlined]{algorithm2e}

\definecolor{cvprblue}{rgb}{0.21,0.49,0.74}
\usepackage[breaklinks,colorlinks,allcolors=cvprblue]{hyperref}

\title{TC-IDM: Grounding Video Generation for Executable \\ Zero-shot Robot Motion}

\author{
\small
\vspace{-2em}
\\
Weishi Mi\textsuperscript{\rm 1$^{\dagger}$}, 
Yong Bao\textsuperscript{\rm 1$^{\dagger}$},
Xiaowei Chi\textsuperscript{\rm 1,2,3,$^{\dagger}$}, 
Xiaozhu Ju\textsuperscript{\rm 1}, 
Zhiyuan Qin\textsuperscript{\rm 1}, \\
Kuangzhi Ge\textsuperscript{\rm 2}, 
Kai Tang\textsuperscript{\rm 1,2},
Peidong Jia\textsuperscript{\rm 1,2},
Shanghang Zhang~\textsuperscript{2,\Envelope}, Jian Tang~\textsuperscript{1,\Envelope} \\
\small\textsuperscript{\rm 1} Beijing Innovation Center of Humanoid Robotics, \\
\small\textsuperscript{\rm 2} State Key Laboratory of Multimedia Information Processing, School of Computer Science, Peking University, \\
\small\textsuperscript{\rm 3} Hong Kong University of Science and Technology \\
}

\date{}  

\begin{document}
\maketitle

\begin{abstract}
The vision–language–action (VLA) paradigm has enabled powerful robotic control by leveraging vision–language models, but its reliance on large-scale, high-quality robot data limits its generalization. Generative world models offer a promising alternative for general-purpose embodied AI, yet a critical gap remains between their pixel-level plans and physically executable actions.
To this end, we propose the \textbf{Tool-Centric Inverse Dynamics Model (TC-IDM)}. By focusing on the tool's imagined trajectory as synthesized by the world model, TC-IDM establishes a robust intermediate representation that bridges the gap between visual planning and physical control.
TC-IDM extracts the tool's point cloud trajectories via segmentation and 3D motion estimation from generated videos. Considering diverse tool attributes, our architecture employs decoupled action heads to project these planned trajectories into 6-DoF end-effector motions and corresponding control signals.
This 'plan-and-translate' paradigm not only supports a wide range of end-effectors but also significantly improves viewpoint invariance. Furthermore, it exhibits strong generalization capabilities across long-horizon and out-of-distribution tasks, including interacting with deformable objects.
In real-world evaluations, the world model with TC-IDM achieves an average success rate of \textbf{61.11\%}, with \textbf{77.7\%} on simple tasks, and \textbf{38.46\%} on zero-shot deformable object tasks—substantially outperforming end-to-end VLA-style baselines and other IDMs.
\vspace{1em} 
\noindent
\begin{center}
    \textbf{Project Page:} \href{https://wsbaiyi.github.io/TC-IDM-Page/}{wsbaiyi.github.io/TC-IDM-Page} \\
    \vspace{0.3em} 
    \textbf{Code:} \href{https://github.com/wsbaiyi/TC-IDM}{github.com/wsbaiyi/TC-IDM}
\end{center}


\end{abstract}    
\section{Introduction}
\label{sec:intro}

\begin{figure}[h]
    \centering
    \includegraphics[width=\linewidth]{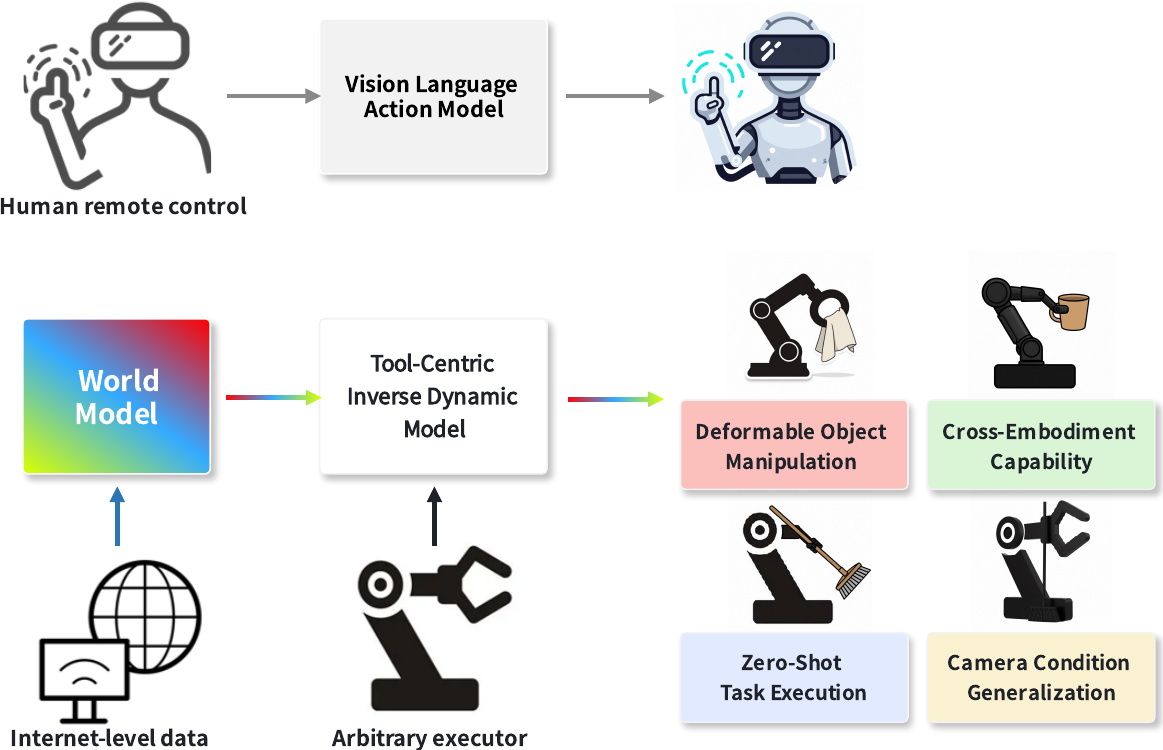}
    \caption{Tool-centric world-model-based IDM achieves stronger generalization than VLA baselines, providing a clean and rigid structure that is easier to learn and transfer across embodiments.}
    \label{fig:teaser}
    \vspace{-4mm}
\end{figure}

With the introduction of end-to-end vision–language–action models such as OpenVLA, $\pi_{0}$, the RT series, and Octo~\citep{kim2024openvla,black2024pi0visionlanguageactionflowmodel,zitkovich2023rt,team2024octo}, transferring the strong perceptual and reasoning capabilities of vision–language models into embodied settings has become feasible. These approaches typically rely on a pretrained base model obtained from large-scale Visual-Language-Action (VLA) datasets, which is subsequently trained or fine-tuned on the specific manipulation data. The success of this paradigm hinges on access to extensive high-quality robotic datasets~\citep{walke2024bridgedatav2datasetrobot, fu2024mobilealohalearningbimanual,embodimentcollaboration2025openxembodimentroboticlearning}, making it difficult for such models to generalize to out-of-distribution (OOD) tasks and environments. 


To advance toward general-purpose embodied intelligence, emerging paradigms leverage generative models as high-level planners to provide "visual foresight." In this framework, predicted future states are translated into executable robotic commands via a downstream policy module~\citep{yang2023unisim,hu2025videopredictionpolicygeneralist,chi2025mindlearningdualsystemworld}. Recent breakthroughs in video-based world models, exemplified by Sora~\citep{videoworldsimulators2024}, Kling~\cite{Kling}, Cosmos~\citep{cosmos2}, and WoW~\citep{chi2025wowworldomniscientworld}, have demonstrated an unprecedented capacity to synthesize high-fidelity, physically plausible video sequences. Such "visual foresight" allows agents to internalize the intricate dynamics required for complex manipulation, from precision grasping to non-rigid folding. By simulating potential future states in pixel space, these models serve as a powerful generative planning prior, addressing long-standing bottleneck tasks in robotics that were previously considered computationally or physically intractable.

Despite these advances, a persistent misalignment exists between the visual planning and low-level control. This planning-action chasm represents a formidable 'last-mile' challenge, precluding the direct deployment of visual planners on physical hardware. The raw output of a world model, which is a sequence of RGB frames, does not inherently constitute a set of robotic control primitives. Such visual synthesis often suffers from noisy trajectories, physical hallucinations, or kinematic inconsistencies, largely because it lacks a grounded representation of the robot’s own morphology and actuation limits. Consequently, the primary challenge lies in the robust translation of these high-level pixel-based plans into low-level, physically consistent trajectories within the joint-space or torque-control domain. Methods that attempt to track object states from videos, such as AVDC~\citep{ko2023learning}, VidBot~\citep{chen2025vidbot}, and Novaflow~\citep{li2025novaflow}, are often brittle, failing when faced with occlusions, rapid motion, or deformable objects like cloth, whose state is notoriously difficult to define and track.

To address these limitations, we introduce \textbf{Tool-Centric Inverse Dynamics Model (TC-IDM)}, as illustrated in Fig.~\ref{fig:teaser}. Our core insight illuminates a different path: instead of relying on the unstable state of the \emph{object or full pixels}, we anchor our control policy to the stable, well-defined motion of the \emph{robot's end-effector} as envisioned by the world model. The tool's trajectory, as imagined in the generated video, serves as a robust and direct intermediate representation for linking high-level visual planning with low-level physical control. Our "plan-and-translate" pipeline operates in two stages. First, given a video generated by a world model, TC-IDM first uses modern segmentation models, such as Segment Anything Model 3 (SAM 3)~\citep{carion2025sam3segmentconcepts} and a 3D motion estimator to accurately recover a set of 6-DoF target trajectories for a dense collection of points on the end-effector from the generated video. Then, two learned action prediction heads respectively transform these tool-centric trajectories and the fused visual–textual command features into executable robot manipulation trajectories and tool control signals.

This decoupled design grants our system several key advantages. It is inherently robust to camera viewpoint variations because the 6-DoF tool trajectory is represented in a world coordinate frame. It is flexible enough for deformable object manipulation, as we no longer need to model the complex state of the cloth, only the tool's intended path. Finally, it is adaptive to diverse, long-horizon tasks by leveraging the compositional planning of world models while ensuring the feasibility of each sub-action. In real-world evaluations, the world model with TC-IDM achieves an average success rate of \textbf{61.11\%}, with \textbf{77.7\%} on simple tasks, and \textbf{38.46\%} on zero-shot deformable object tasks—substantially outperforming end-to-end VLA-style baselines and other inverse dynamic models. Our core contributions are:
\begin{itemize}
    \item The proposal of TC-IDM, a novel framework that bridges world model planning and robot control by using the tool’s trajectory as a key intermediate representation.
    \item By training separate arm and tool policy heads driven respectively by physical motion cues and semantic visual features, TC-IDM cleanly decouples and effectively exploits heterogeneous information streams.
    \item A comprehensive set of real-world experiments that validate our method's effectiveness and establish a new, strong baseline for deploying generative world models on physical robots.
\end{itemize}
\section{Related Work}
\label{sec:related_work}

\paragraph{Video-conditioned Inverse Dynamics Models for Robotic Manipulation.} Recent work investigates video-conditioned inverse dynamics models (IDMs) as a bridge between visual prediction and actionable robot control. VidBot~\citep{chen2025vidbot} extracts object-level 3D affordances and interaction trajectories from in-the-wild human videos to enable zero-shot manipulation. AVDC~\citep{ko2023learning} shows that dense correspondences between synthesized future frames can supervise action inference without labels, allowing policies to act purely from predicted object motion. AnyPos~\citep{tan2025anypos} complements these efforts with a task-agnostic IDM trained on large-scale random exploration, using a video-conditioned validation module to maintain feasibility during deployment. Collectively, recent works demonstrate a trend toward leveraging generative video models and image-space correspondences as general supervisory signals, reducing reliance on task-specific demonstrations.

\paragraph{End-to-End Vision-Language-Action (VLA) Models.} End-to-end VLA models such as RT-2~\citep{zitkovich2023rt}, OpenVLA~\citep{kim2024openvla}, and diffusion-based policies~\citep{chi2025diffusion} map multimodal inputs directly to low-level actions and perform well in-distribution, but struggle with long-horizon and compositional reasoning due to the lack of explicit planning. Data augmentation methods like DreamVLA~\citep{zhang2025dreamvla} offer limited improvement. To bridge this gap, recent work introduces intermediate structures: hierarchical models like $\pi_{0.5}$~\citep{intelligence2504pi0} decompose tasks into subtasks, DexGraspVLA~\citep{zhong2025dexgraspvla} uses visual affordance cues, and MolmoAct~\citep{lee2025molmoact} and GraspVLA~\citep{deng2025graspvla} embed reasoning via spatial and pose predictions. These approaches show that explicit intermediate reasoning improves long-horizon control.

\begin{figure*}[h]
  \centering
  \includegraphics[width=\textwidth]{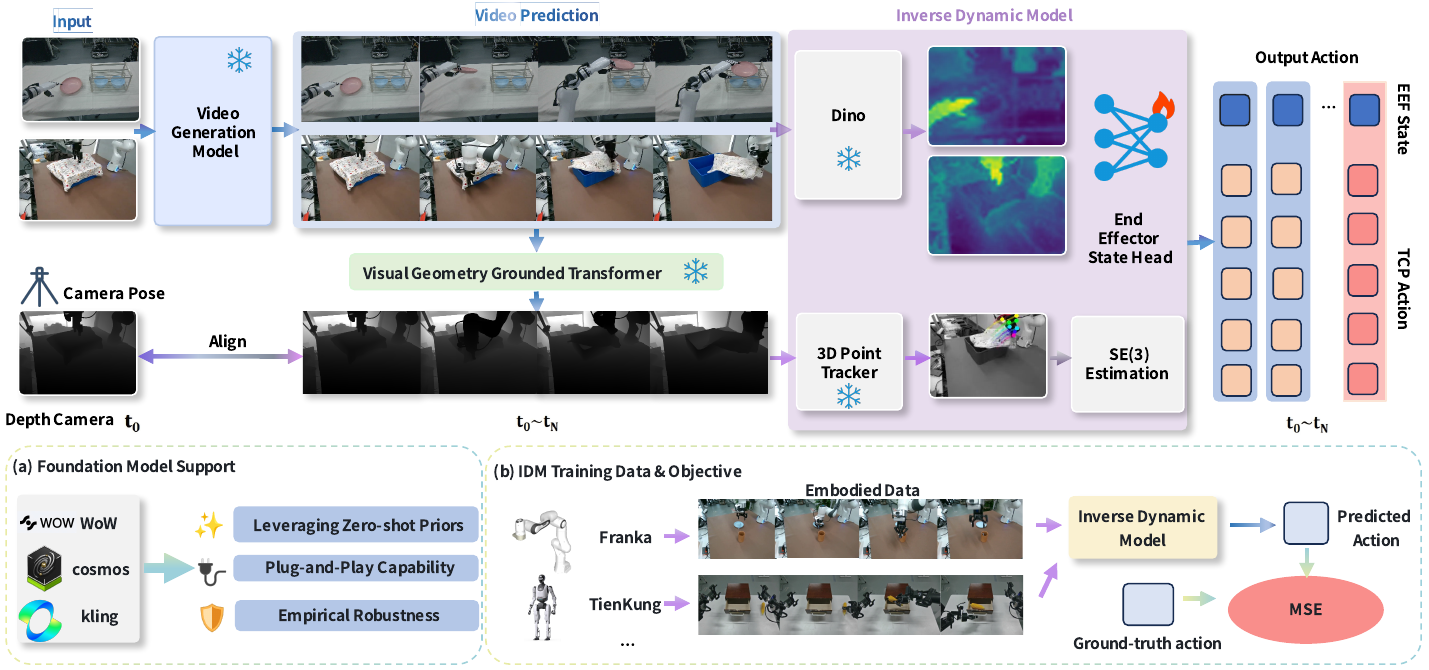} 
  \caption{\textbf{Overview of Our Tool-Centric Video Generation Framework} Given an initial RGB frame, depth map, and text instruction, our method utilizes a world model to generate high-level planning for extracting semantic vision features (DINOv3~\citep{simeoni2025dinov3} in our case), depth-aligned geometry(using Video Depth Anything~\citep{chen2025video}). We further employ a 3D motion tracker~\citep{xiao2025spatialtrackerv2} to derive gripper-centric flow, guided jointly by the high-level plan and a gripper mask obtained via SAM 3~\citep{carion2025sam3segmentconcepts}. Guided by the semantic vision feature and motion flow, the gripper and gesture states are respectively predicted via dedicated MLP heads.}
  \label{fig:method_main}
  \vspace{-2mm}
\end{figure*}

\paragraph{Generative Models for Visual Planning.} A complementary line of work uses video generative models—especially diffusion-based world models—to implement "plan-as-generation."~\citep{hu2025videopredictionpolicygeneralist,chi2025mindlearningdualsystemworld,jia2025video2actdualsystemvideodiffusion} Systems like UniSim~\citep{yang2023unisim}, EVA~\citep{chi2024eva}, and RoboDreamer~\citep{zhou2024robodreamer} imagine task-achieving future videos that guide downstream controllers. However, the subsequent pixel-to-action translation, often via an IDM, is brittle and sensitive to visual artifacts, occlusions, and non-rigid motion. The frequent trust-region replanning required by methods such as VPA~\citep{wang2019learning} further illustrates the gap between imagined plans and physical execution.


\section{Methodology}
\label{sec:method}

Generative world models provide powerful, visual foresight, but a "last-mile" gap remains between their pixel-level plans and physically executable robot actions. To bridge this, we introduce the Tool-Centric Inverse Dynamics Model (TC-IDM), a framework that translates imagined futures into precise, low-level control.

Our proposed Inverse Dynamics Model focuses on the end-effector by decoupling robotic control into two distinct components: the internal state of the end-effector in Sec.\ref{sssec:vision-semantic-state-generation} and the global 6-DoF pose trajectory of the Tool Center Point in Sec.\ref{sssec:geometry_grounded_gesture}. Importantly, the proposed architecture is agnostic to the choice of end-effector and can be instantiated with arbitrary tool embodiments. In the following, we present the formulation using a two-finger parallel gripper as a concrete example. The overall architecture is illustrated in Fig.~\ref{fig:method_main}.

\subsection{Spatiotemporal   Prediction.}
\label{ssec:synthesis}


\paragraph{Video Prediction.}
A video-generative world model~\citep{chi2025wowworldomniscientworld, nvidia2025cosmosworldfoundationmodel}, conditions on $I_{rgb}^0$ and $L$ to generate an imagined RGB video sequence as the high-level planning, $V_{rgb\_gen} = \{I_{rgb\_gen}^t\}_{t=0}^T$. In the first frame, it is essential to ensure that the robot’s end-effector is visible within the field of view. This enables the acquisition of its initial pose and helps prevent potential collisions during subsequent motion planning. 


\paragraph{Spatial Estimation}
Given the video sequence, we use VGGT~\cite{wang2025vggt} to generate an initial depth map sequence \(\{\tilde{D}_t\}_{t=1}^{T}\) and corresponding relative camera poses \(\{\mathbf{P}_t^{\text{rel}}\}_{t=1}^{T}\). A reference metric depth map \(D_{\text{metric}}^0\) is obtained by refining a raw depth measurement with a completion network~\cite{liu2025manipulation}.

We then compute a scale \(s\) and shift \(d\) by linearly aligning the first predicted depth map \(\tilde{D}_0\) to \(D_{\text{metric}}^0\) via least squares:
\begin{equation}
s, d = \arg\min_{s, d} \sum_{i \in \Omega} \left\| s \cdot \tilde{D}_0(\mathbf{p}_i) + d - D_{\text{metric}}^0(\mathbf{p}_i) \right\|^2.
\end{equation}
These parameters are applied to all depth maps: \(D_t^{\text{metric}} = s \cdot \tilde{D}_t + d\). Using the known camera pose \(\mathbf{P}_{\text{gt}}^0\) of the reference frame, we solve for a rigid transformation \((\mathbf{R}, \mathbf{t})\) that aligns the scaled relative pose \(s \cdot \mathbf{P}_0^{\text{rel}}\) to \(\mathbf{P}_{\text{gt}}^0\). This transformation is applied to all poses:
\[
\mathbf{P}_t^{\text{metric}} = \begin{bmatrix} \mathbf{R} & \mathbf{t} \end{bmatrix} \cdot (s \cdot \mathbf{P}_t^{\text{rel}}),
\]
resulting in a fully metric camera trajectory \(\{\mathbf{P}_t^{\text{metric}}\}_{t=1}^{T}\) and depth sequence \(\{D_t^{\text{metric}}\}_{t=1}^{T}\).




\subsection{Decoupled Action Translation}
\label{ssec:translation}

Building upon the spatiotemporal prediction obtained in the previous stage, TC-IDM explicitly decomposes action translation into two complementary and decoupled streams: vision-driven state generation and geometry-grounded gesture generation. These two streams are subsequently concatenated to form the final robot control vector, which comprises the gripper action \(A_{\text{gripper}}\) and the end-effector action \(A_{\text{TCP}}\).

\subsubsection{Vision-Driven State Generation}
\label{sssec:vision-semantic-state-generation}

We adopt an \textbf{encoder--decoder architecture} in which a frozen visual encoder extracts temporally-consistent semantic features from generated RGB videos, and a lightweight task-conditioned decoder predicts discrete gripper commands from these features.

\paragraph{Semantic Feature Extraction.}
To extract high-level task semantics, the generated RGB video $V_{\text{rgb-gen}} = \{I^t\}_{t=0}^T$ is processed by a frozen, pre-trained visual encoder  \textbf{DINOv3}~\citep{simeoni2025dinov3} with strong dense correspondence capabilities. The encoder produces a sequence of semantic embeddings
\begin{equation}
F_{\text{dino}} = \{f_{\text{dino}}^t\}_{t=0}^T,
\end{equation}
which capture task-critical cues such as object identity, contact states, and phase transitions.

\paragraph{Gripper MLP Head (1-DoF Gripper Control).}
Based on the extracted semantic feature sequence, the gripper control network predicts a continuous grasping aperture parameter via a lightweight MLP: 
\begin{equation}
\label{eq:gripper_head_cvpr}
A{_\text{gripper}} = \text{GripperHead}\left({f_{\text{dino}}^t} \right).
\end{equation}

\subsubsection{Geometry-Grounded Gesture Generation}
\label{sssec:geometry_grounded_gesture}

The action space of the robotic gesture is defined as the 6D pose of the Tool Center Point (TCP), a fixed reference point attached to the robot's end-effector. This branch operates by performing explicit 3D motion tracking of the rigid-body transformation of the target part, as predicted by the aligned spatiotemporal prediction. Using this geometric information, the required end-effector trajectory is derived analytically through rigid-body kinematics. The approach does not rely on learned action decoders, thereby maintaining interpretability in motion generation. 

\paragraph{Dense Gripper Trajectory Extraction.}
As input, we utilize the aligned spatiotemporal prediction
$V_{\text{spatial}} = \{I_{rgb\_gen}^t, D_{\text{metric}}^t\}_{t=0}^T$,
produced by the  in Sec.~\ref{ssec:synthesis}.
To isolate the end-effector, we generate a dense gripper mask $M_{\text{gripper}}^t$ for each frame using \textbf{SAM3}~\citep{carion2025sam3segmentconcepts}. The mask is used to suppress background motion and constrain tracking to the gripper region. Conditioned on the aligned RGB-D input and the gripper mask, we employ 3D Point Tracker~\citep{xiao2025spatialtrackerv2} to extract dense 3D point trajectories across time:
\begin{equation}
\label{eq:spatialtracker}
\mathcal{T}_{\text{dense}}
=
\text{PointTracker}\!\left(
V_{\text{spatial}}, M_{\text{gripper}}
\right),
\end{equation}
where each trajectory $\{\mathbf{x}_i^t\}_{t=0}^{T}$ represents the temporally tracked 3D motion of a gripper surface point in an robot base frame. Unlike static lifting, SpatialTrackerv2~\citep{xiao2025spatialtrackerv2} explicitly enforces temporal correspondence and motion coherence, which is critical for reliable downstream pose recovery.

Although the spatial prediction is metrically aligned, generative artifacts may still introduce locally inconsistent tracks. To obtain a stable geometric representation, we further enforce a rigid-body motion prior. Specifically, we evaluate the consistency of each dense trajectory under a rigid transformation and retain the top-$K$ trajectories that best satisfy this assumption, with $K=10$ in all experiments:
\begin{equation}
\label{eq:rigid_filter}
\mathcal{T}_{\text{gripper}}
=
\text{Select}_{\text{rigid}}\!\left(\mathcal{T}_{\text{dense}}\right).
\end{equation}

\paragraph{Rigid-Body Pose Recovery for 6-DoF Control.}
Rather than learning an action decoder, we recover executable robot actions through explicit rigid-body alignment. Assuming locally rigid gripper motion between consecutive frames, we estimate the relative pose transformation that best aligns the filtered 3D trajectories.

At timestep $t$, given $\mathcal{T}_{\text{gripper}}^t = \{\mathbf{x}_i^t\}_{i=1}^{K}$ and
$\mathcal{T}_{\text{gripper}}^{t+1}$, we solve for the rigid-body transformation $(\mathbf{R}_t, \mathbf{p}_t) \in SE(3)$:
\begin{equation}
\label{eq:rigid_alignment}
(\mathbf{R}_t, \mathbf{p}_t)
=
\arg\min_{\mathbf{R} \in SO(3),\, \mathbf{p} \in \mathbb{R}^3}
\sum_{i=1}^{K}
\left\|
\mathbf{R}\,\mathbf{x}_i^t + \mathbf{p} - \mathbf{x}_i^{t+1}
\right\|^2.
\end{equation}

The recovered transformation directly defines the 6-DoF end-effector action executed at timestep $t+1$:
\begin{equation}
\label{eq:6dof_action}
\text{A}_{TCP}^{t+1}
=
(\mathbf{R}_t, \mathbf{p}_t).
\end{equation}

By grounding control in temporally consistent 3D trajectories extracted by 3D motion tracker and enforcing analytic rigid-body constraints, this branch ensures faithful and physically interpretable execution of the imagined scene dynamics.

\begin{figure*}[h]
    \centering
    \includegraphics[width=\linewidth]{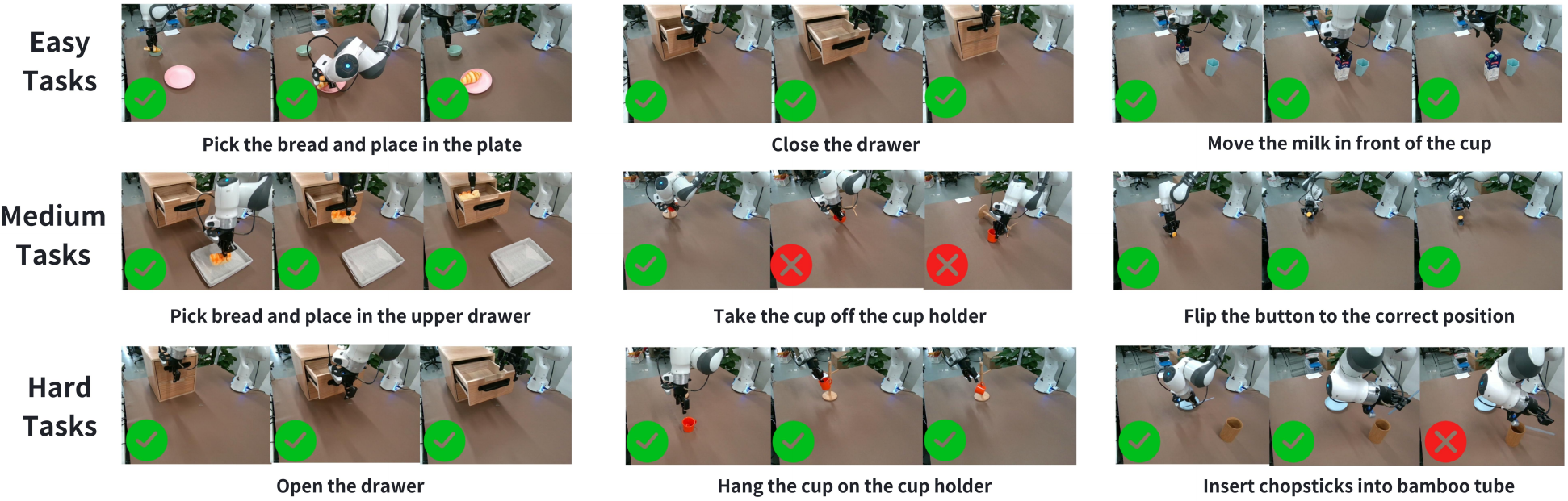}
    \caption{\textbf{Representative manipulation tasks categorized by difficulty.} The figure shows qualitative examples of tasks executed by the robot, grouped into three levels: Easy (top row), Medium (middle row), and Hard (bottom row). Easy tasks involve basic pick-and-place or push actions (e.g., placing bread on a plate); Medium tasks require more precise control or obstacle avoidance (e.g., taking a cup from a holder); Hard tasks involve fine-grained 6-DoF coordination (e.g., inserting chopsticks into a bamboo tube). This categorization provides a structured benchmark for evaluating the model's performance across spatial and temporal complexity.
    }
    \label{fig:Evaluation_of_Real-World_Replay}
    \vspace{-4mm}
\end{figure*}

\begin{table*}[h!]
\centering
\caption{Comparison of Success Rate (\%) on Easy, Medium and Hard tasks across different models on ground truth video replay.}
\label{tab:relpay_results}
\resizebox{\linewidth}{!}{%
\begin{tabular}{@{\hskip 5pt}l@{\hskip 5pt} | @{\hskip 5pt}ccc@{\hskip 5pt}| @{\hskip 5pt}ccc@{\hskip 5pt} | @{\hskip 5pt}ccc@{\hskip 5pt}}
\toprule
\rowcolor{gray!15}
\multirow{2}{*}{\textbf{Model / Task}} & \multicolumn{3}{c|}{\textbf{Easy Tasks}} & \multicolumn{3}{c|}{\textbf{Medium Tasks}} & \multicolumn{3}{c}{\textbf{Hard Tasks}} \\
\cmidrule(lr){2-7} \cmidrule(lr){8-10}
\rowcolor{gray!10}
 & \textbf{Bread to Plate} & \textbf{Close Drawer} & \textbf{Move Milk} & \textbf{Bread in Drawer} & \textbf{Take Cup} & \textbf{Open Drawer} & \textbf{Flip Button} & \textbf{Hang Cup} & \textbf{Insert Chopsticks} \\
\midrule
ResNet-MLPs~\citep{he2016deep}     & 6/10 & 7/10 & 5/10 & 4/10 & 1/10 & 2/10 & 0/15 & 0/15 & 1/15 \\
AVDC~\citep{ko2023learning}        & 3/10 & 4/10 & 4/10 & 2/10 & 1/10 & 1/10 & 0/15 & 0/15 & 0/15 \\
Anypos~\citep{tan2025anypos}        & 8/10 & 9/10 & 8/10 & 7/10 & 6/10 & 3/10 & 2/15 & 2/15 & 0/15 \\
2DtrackerIDM        & 8/10 & 10/10 & 8/10 & 8/10 & 5/10 & 4/10 & 3/15 & 3/15 & 1/15 \\
\rowcolor{blue!10}
\textbf{TC-IDM (Ours)} & 10/10 & 9/10 & 9/10 & 8/10 & 4/10 & 4/10 & 2/15 & 7/15 & 4/15 \\
\bottomrule
\end{tabular}
}
\end{table*}

\begin{table*}[h!]
\centering
\caption{Comparison of Success Rate (\%) on Easy, Medium, and Hard tasks across different video generation models. Models marked with * are fine-tuned using embodied data.}
\label{tab:vla_results}
\resizebox{\linewidth}{!}{%
\begin{tabular}{@{\hskip 5pt}l@{\hskip 5pt} |  @{\hskip 5pt}ccc@{\hskip 5pt}| @{\hskip 5pt}ccc@{\hskip 5pt} | @{\hskip 5pt}ccc@{\hskip 5pt}}
\toprule
\rowcolor{gray!15}
\multirow{2}{*}{\textbf{Model / Task}} & \multicolumn{3}{c|}{\textbf{Easy Tasks}} & \multicolumn{3}{c|}{\textbf{Medium Tasks}} & \multicolumn{3}{c}{\textbf{Hard Tasks}} \\
\cmidrule(lr){2-7} \cmidrule(lr){8-10}
\rowcolor{gray!10}
 & \textbf{Bread to Plate} & \textbf{Close Drawer} & \textbf{Move Milk} & \textbf{Bread in Drawer} & \textbf{Take Cup} & \textbf{Open Drawer} & \textbf{Flip Button} & \textbf{Hang Cup} & \textbf{Insert Chopsticks} \\
\midrule
\rowcolor{gray!10}
$\pi_{0}$~\citep{black2024pi_0}       & 2/9 & 4/9 & 2/9 & 0/9 & 0/9 & 0/9 & 0/9 & 0/9 & 0/9 \\
CogVideo-IDM~\citep{hong2022cogvideo} & 0/9 & 0/9 & 0/9 & 0/9 & 0/9 & 0/9 & 0/9 & 0/9 & 0/9 \\
wan2.1-IDM~\citep{wan2025wan} & 0/9 & 0/9 & 0/9 & 0/9 & 0/9 & 0/9 & 0/9 & 0/9 & 0/9 \\
cosmos-1-IDM~\citep{agarwal2025cosmos} & 0/9 & 0/9 & 0/9 & 0/9 & 0/9 & 0/9 & 0/9 & 0/9 & 0/9 \\
Hailuo-IDM~\citep{li2025minimax} & 1/9 & 1/9 & 0/9 & 0/9 & 0/9 & 0/9 & 0/9 & 0/9 & 0/9 \\
Kling-IDM~\citep{Kling} & 3/9 & 0/9 & 5/9 & 0/9 & 0/9 & 0/9 & 0/9 & 0/9 & 0/9 \\
Cosmos2-IDM~\citep{cosmos2} & 4/9 & 2/9 & 0/9 & 1/9 & 0/9 & 0/9 & 0/9 & 0/9 & 0/9 \\

WoW-cosmos2-IDM*~\citep{chi2025wowworldomniscientworld} & 8/9 & 3/9 & 1/9 & 3/9 & 0/9 & 0/9 & 0/9 & 0/9 & 0/9 \\
\textbf{WoW-wan-IDM*~\citep{chi2025wowworldomniscientworld}} & 9/9 & 5/9 & 7/9 & 5/9 & 3/9 & 2/9 & 0/9 & 1/9 & 1/9 \\
\bottomrule
\end{tabular}
}
\end{table*}

\section{Quantitative Manipulation Experiments}

\subsection{Quantitative Experiment Setups}

\paragraph{Task Difficulty Classification}
In the process of implementation, we identified four common failure modes: (1) Unintended collisions during multi-DoF control; (2) Substantial inaccuracies in rotational movements; (3) Insufficient translational precision; and (4) Incorrect gripper control. Based on these observations, we classify tasks into three tiers based on DoF and precision constraints:
\begin{itemize}
    \item \textbf{Hard Tasks:} Require at least 5-DoF control or tolerate errors below $2~cm/10^{\circ}$.
    \item \textbf{Medium Tasks:} Require at least 4-DoF control or involve simple collision avoidance.
    \item \textbf{Easy Tasks:} All remaining tasks.
\end{itemize}

\paragraph{Representative Task Examples}
Fig.~\ref{fig:Evaluation_of_Real-World_Replay} provides qualitative examples of these tasks, which we categorize according to our defined difficulty scheme.
\begin{enumerate}
    \item \textbf{Pick the bread and place in the plate (Easy):} A standard "pick-and-place" task.
    \item \textbf{Close the drawer (Easy):} The robot performs a "push" action to close the drawer.
    \item \textbf{Move the milk in front of the cup (Easy):} A task requiring  spatial positioning.
    \item \textbf{Pick bread and place in the upper drawer (Medium):} A compound task combining pick-and-place with vertical obstacle avoidance.
    \item \textbf{Take the cup off the cup holder (Medium):} A standard grasping task from a constrained holder.
    \item \textbf{Open the drawer (Medium):} The robot needs to precisely insert its gripper into the gap between the handle and the cabinet body.
    \item \textbf{Flip the button to the correct position (Hard):} A task demanding precise rotation to achieve the button's final vertical alignment.
    \item \textbf{Hang the cup on the cup holder (Hard):} A high-precision 6-DoF placement task requiring precise alignment of the cup handle.
    \item \textbf{Insert chopsticks into bamboo tube (Hard):} A challenging "insertion" task requiring high translational and rotational precision.
\end{enumerate}

\subsection{Manipulation Evaluation of Real-World Replay}
Table~\ref{tab:relpay_results} presents quantitative results on real-world replay across Easy, Medium, and Hard manipulation tasks using different inverse dynamics models and perception modules. The comparison reveals several consistent trends.

\textbf{Inverse Dynamics Models.}
AnyPos~\citep{tan2025anypos} substantially outperforms AVDC~\citep{ko2023learning} across all difficulty levels. While both models achieve moderate success on Easy tasks, AVDC exhibits pronounced temporal drift, leading to compounding errors on long-horizon interactions such as \emph{Bread in Drawer} and \emph{Open Drawer}. This discrepancy becomes critical on Hard tasks, where AVDC fails almost entirely, whereas AnyPos maintains non-trivial performance on skills requiring precise multi-step control (e.g., \emph{Flip Button}, \emph{Hang Cup}). These results highlight that the stability of the inverse dynamics model is a dominant factor governing replay reliability.

\textbf{Perception Ablations.}
The ablation on motion perception demonstrates that spatial encoders alone are insufficient for accurate inverse dynamics estimation. The ResNet~\citep{he2016deep} baseline yields limited success, indicating that static appearance cues do not adequately capture action-relevant motion structure. Incorporating a 2D tracker, cotracker3~\citep{karaev2024cotracker} improves performance through explicit pixel correspondence, yet the absence of geometric or depth-aware reasoning constrains its effectiveness on Hard tasks.

Our proposed TC-IDM achieves the highest success rate across nearly all tasks. It attains perfect performance on Easy tasks and outperforms all baselines on Medium tasks. Notably, on the most challenging settings—requiring fine-grained coordination and long-horizon consistency— TC-IDM significantly surpasses all alternatives. It achieves \textbf{93.3\% on easy tasks}, \textbf{53.3\% on medium tasks}, and \textbf{28.9\% on hard tasks}. We note that 93.3\% action replay accuracy represents the upper bound for task success of the IDM in this evaluation.

\subsection{Manipulation Evaluation across Video World Models}

\begin{figure*}[h]
    \centering
    \includegraphics[width=0.9\linewidth]{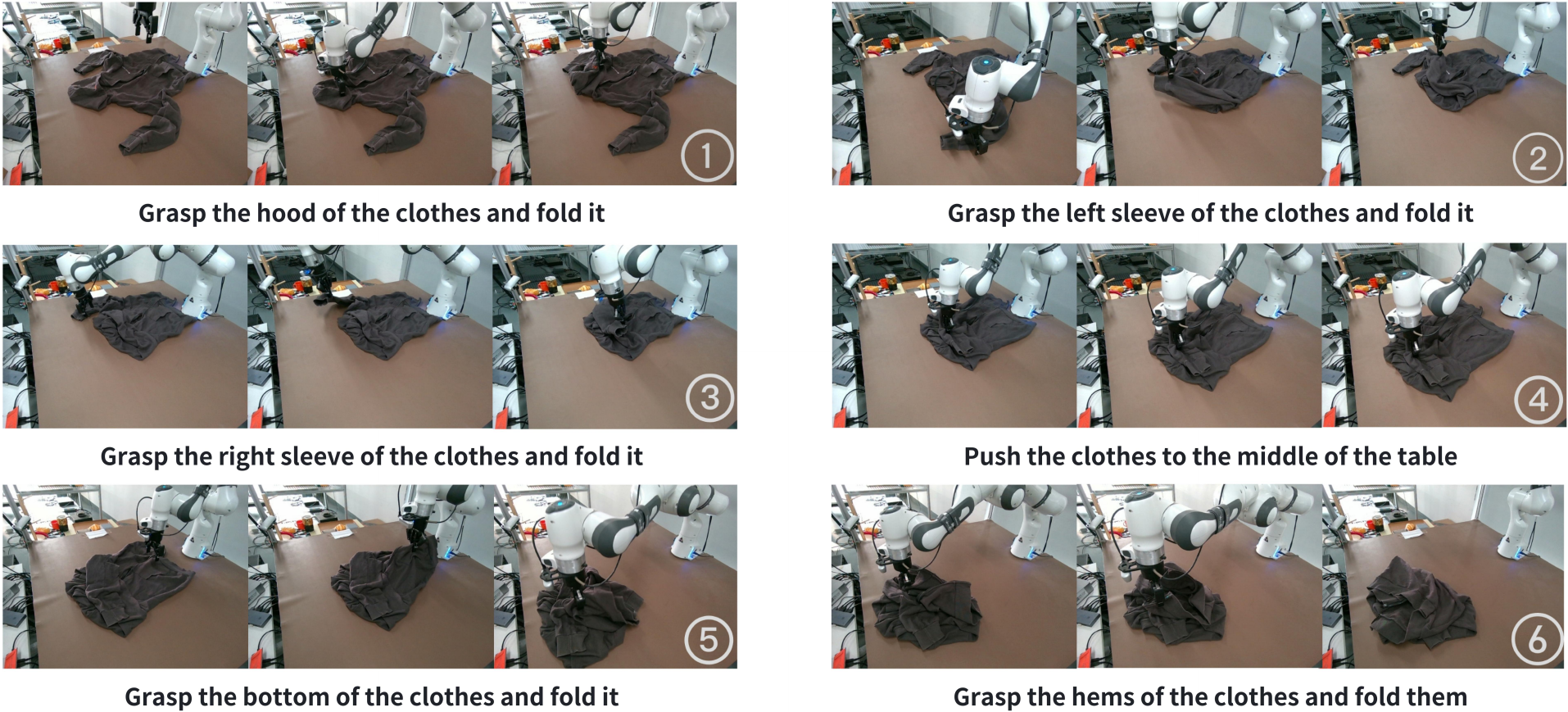}
    \caption{\textbf{Long-Horizon Generalization.}The world model generates a full six-step visual plan from the initial scene and the instruction “Fold the hoodie,” and the TC-IDM executes the plan in the real world, completing all stages of the folding sequence.
    }
    \label{fig:long_horizon_task}
    \vspace{-4mm}
\end{figure*}

We evaluate manipulation performance across multiple world models and their IDMs, including {$\pi_{0}$}~\citep{black2024pi_0}, {CogVideo-IDM}~\citep{hong2022cogvideo} /{cosmos-1-IDM}~\citep{agarwal2025cosmos} /{Wan2.1-IDM}~\citep{wan2025wan}, {Hailuo-IDM}~\citep{li2025minimax}, {Kling-IDM}~\citep{Kling}, {Cosmos2-IDM}~\citep{cosmos2}, {WoW-cosmos2-IDM}~\citep{chi2025wowworldomniscientworld} and {WoW-wan-IDM}~\citep{chi2025wowworldomniscientworld}. Table~\ref{tab:vla_results} reports the success rates.

Overall, Our method consistently delivers the most stable and coherent multi-step motion predictions across different world models, with particularly pronounced advantages on complex long-horizon and fine-manipulation tasks; additional visualizations are provided in the appendix.

Quantitative comparisons in Table~\ref{tab:vla_results} confirm these observations:
our method consistently achieves the highest success rate across all difficulty levels and
all three video models, and the margin is especially large on Hard tasks. These results
demonstrate that our approach is robust to variations across video generative models and
generalizes effectively even when the visual dynamics differ substantially between them.
Statistically, the world model with TC-IDM achieves an average success rate of 61.11\%, with 77.7\% on simple tasks.




\section{Generalization Experiments}

In this section, we further designed a zero-shot experiment to further demonstrate the model's generalization capabilities across five critical dimensions: Error Range Analysis, Generalization Across Camera Variations, Generalization for Deformable Objects, Long-Horizon Generalization, and Cross-Embodiment Experiment.

\subsection{Error Range Analysis}

\paragraph{Experimental Setting}
We evaluate the precision of plan-to-action translation using a high-accuracy dart-placing task. All methods receive the \textbf{same generated visual plan}—an idealized execution synthesized by the world model—and must reproduce it in the real world. The setup isolates the inverse dynamics quality by controlling the planning component.

\begin{figure}[h]
    \centering
    \includegraphics[width=\linewidth]{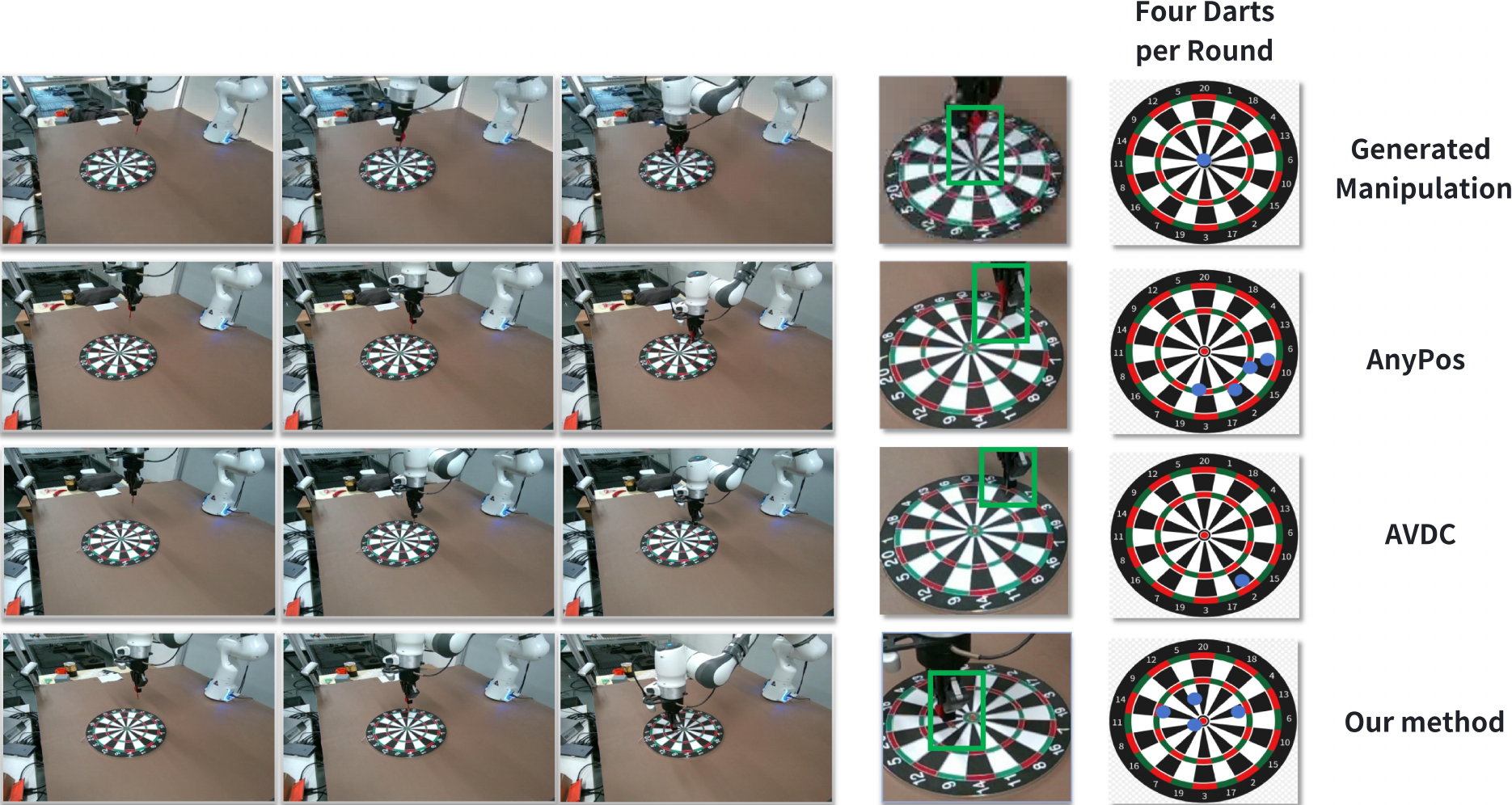}
    \caption{\textbf{Error Range Analysis.}
    All methods receive the same world model-generated manipulation (top) and reproduce the action of placing the dart at the bullseye. The figure compares the execution accuracy of AVDC\cite{ko2023learning}, AnyPos \cite{tan2025anypos}, and our TC-IDM.}
    \label{fig:error_analysis}
    \vspace{-3mm}
\end{figure}

\paragraph{Experimental Analysis}
As shown in Fig.~\ref{fig:error_analysis}, our TC-IDM achieves markedly higher plan-to-execution fidelity than other IDM baselines. Compared to the VLA baseline, the VLA model struggles to follow language instructions and consistently fails to place the dart on the target, exhibiting large deviations from the intended goal. In contrast, TC-IDM reliably throws the dart with high accuracy, consistently hitting within 4 cm of the target. Moreover, by leveraging an Internet-pretrained world model as the trajectory generator, our approach shows strong zero-shot execution capability for diverse manipulation behaviors, whereas VLA models depend heavily on collecting task-specific demonstrations to perform reliably.

\subsection{Generalization Across Camera Variations}
\begin{figure}[h]
    \centering
    \includegraphics[width=\linewidth]{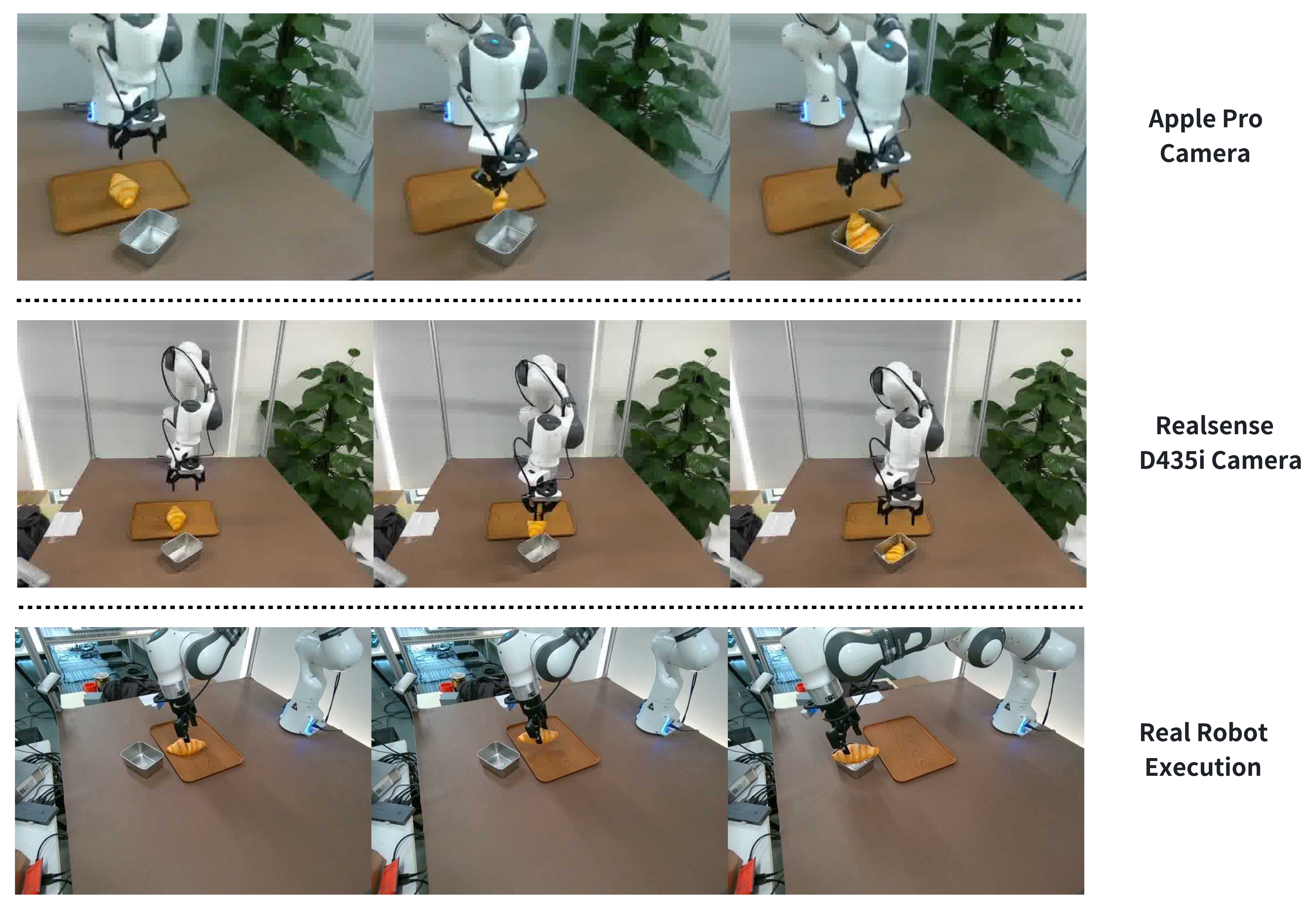}
    \caption{\textbf{Generalization Across Camera Variations.}
    The same task performed using observations from Apple Pro and Realsense D435i.}
    \label{fig:camera_generalization}
    \vspace{-3mm}
\end{figure}

\paragraph{Experimental Setting}
We evaluate whether a model trained on\textit{ Intel Realsense D457 RGB-D observations} can generalize to two unseen camera domains as illustrated in Fig.~\ref{fig:camera_generalization}:
(1) \textit{Apple Pro}, which differs significantly in color tone and exposure;  
(2) \textit{Realsense D435i}, which introduces different viewpoint and noise characteristics.   

\paragraph{Experimental Analysis}
A model trained exclusively on RealSense D457 inputs exhibits strong cross-camera generalization, transferring effectively to both Apple Pro and D435i without any finetuning. Despite substantial variations in color tone, illumination, and viewpoint, the generated trajectories remain highly consistent. Execution results further confirm reliable manipulation across different camera domains, demonstrating the model’s robustness to significant visual distribution shifts.

\subsection{Generalization of Deformable Objects}

\begin{figure}[h]
    \centering
    \includegraphics[width=\linewidth]{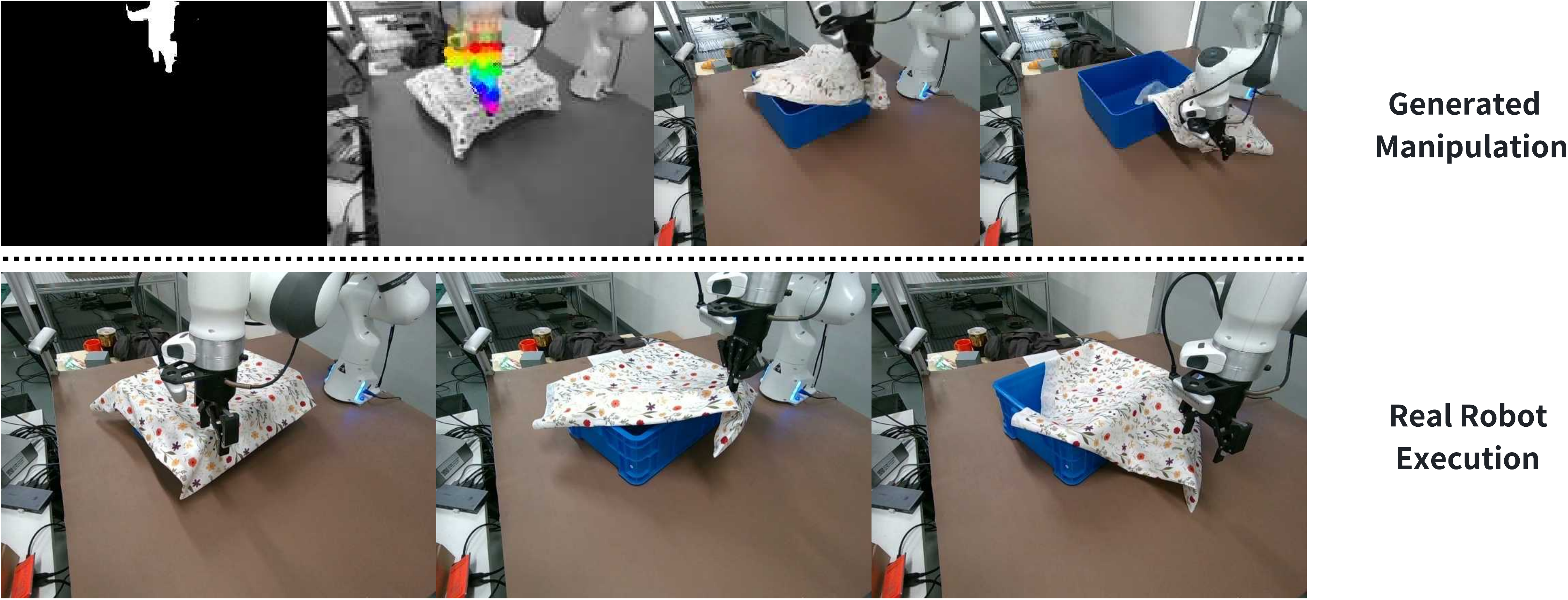}
    \caption{\textbf{Generalization of Deformable Objects.}
    Comparison between generated manipulation (top) and real robot execution (bottom). The model can recognize the cloth as a manipulable object, plan valid grasp points, and generate appropriate removal trajectories.}
    \label{fig:deformable_comparison}
    \vspace{-4mm}
\end{figure}

\paragraph{Experimental Setting}
We perform a zero-shot deformable-object experiment to evaluate how well the model generalizes to non-rigid materials. As shows in Fig.~\ref{fig:deformable_comparison}, the generated visual plan depicting a cloth-removal sequence is provided by the WoW~\citep{chi2025wowworldomniscientworld} world model (top row).
The robot must then reproduce this plan in the real world using TC-IDM, producing the corresponding execution videos (bottom row).  
The cloth’s texture, shape, and folding configuration differ across trials, creating natural variations not seen during training.

\paragraph{Experimental Analysis}
TC-IDM demonstrates zero-shot generalization to deformable-object manipulation, accurately executing the synthesized cloth-handling plan despite being trained exclusively on rigid objects. The model reliably identifies stable grasp regions and predicts robust removal trajectories across diverse cloth geometries and textures, highlighting its ability to generalize grasp and motion prediction beyond its training distribution. Overall, the method achieves a 38.46\% success rate on zero-shot deformable-object tasks.

\subsection{Long-Horizon Generalization}

\paragraph{Experimental Setting}
We evaluate the long-horizon capability of our TC-IDM using a challenging multi-stage cloth-folding task. 
The world model first generates a complete six-step visual plan given the initial scene and the instruction “Fold the hoodie.”  
Our TC-IDM then executes the entire sequence in the real world by translating the generated plan into continuous actions.  
Fig.~\ref{fig:long_horizon_task} shows snapshots of the complete real-robot rollout, illustrating the completion of all six folding stages.

\paragraph{Experimental Analysis}
 The robot can successfully complete the full six-step hoodie-folding procedure in a zero-shot manner, requiring no intermediate corrections or re-planning. Despite substantial state changes as the hoodie deforms after each action, the model maintains coherent multi-step control and preserves close alignment with the intended plan through TC-IDM.

\begin{figure}[h]
\centering
\includegraphics[width=\linewidth]{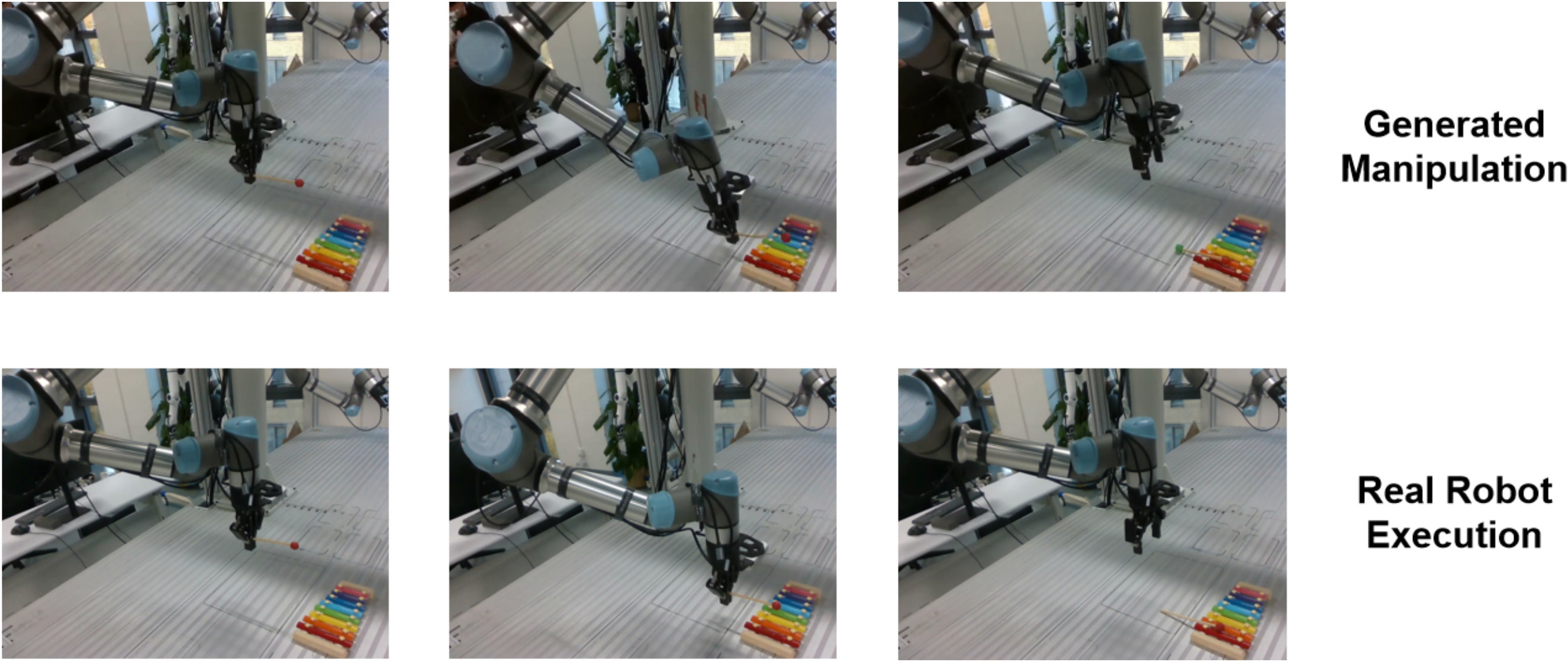}
\caption{\textbf{Cross-Embodiment Experiment.}
Comparison between the generated manipulation (top) and real robot execution (bottom). The model effectively transfers manipulation knowledge to a dual-arm UR robot, enabling precise motion planning and execution for xylophone playing.}
\label{fig:cross_embodiment_xylophone}
\vspace{-4mm}
\end{figure}

\subsection{Cross-Embodiment Experiment}

\paragraph{Experimental Setting}
We evaluate the model's cross-embodiment generalization by transferring motion generation knowledge learned from a single-arm Franka robot to a real dual-arm UR robot. The task requires the robot to play the xylophone by hitting the yellow key with a mallet. We compare the trajectory generated by the model with the execution in the real world, as shown in Fig.~\ref{fig:cross_embodiment_xylophone}.

\paragraph{Experimental Analysis}
We transfer the motion priors learned on a single-arm Franka system effectively to a real dual-arm UR5 platform without any re-training. The executed trajectories closely follow the generated plan, achieving high spatial precision and temporally stable rhythmic motion, indicating the presence of robust embodiment-invariant representations. 

\section{Human-to-Dexterous Hand Migration}
\label{sec:hand_dex}

\begin{figure}[h]
  \centering
  \includegraphics[width=\linewidth]{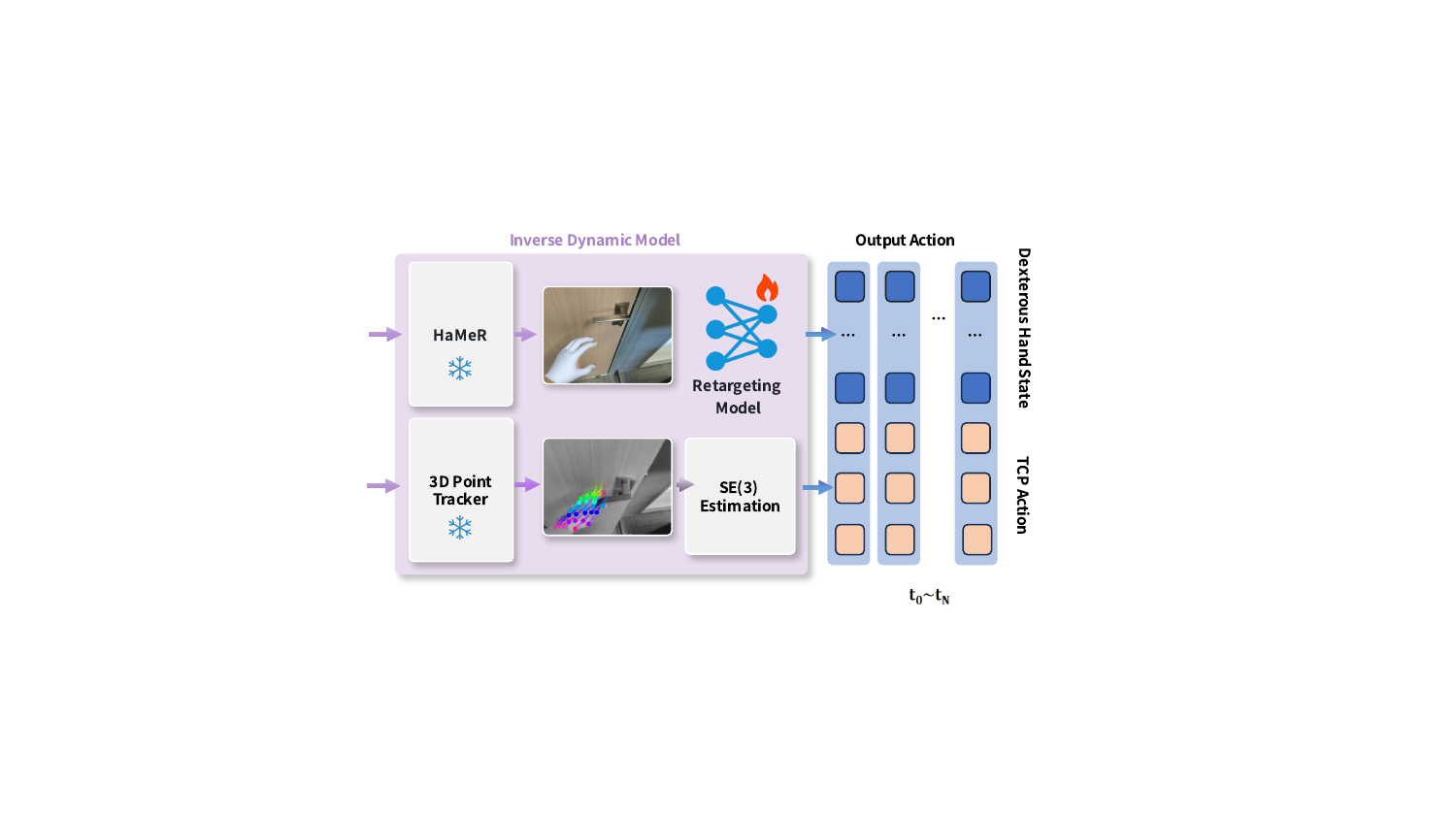} 
  \caption{\textbf{Architecture of the Inverse Dynamic Model for Dexterous Hand Retargeting.} The pipeline consists of two parallel streams: (1) The \textbf{Hand Retargeting Branch} (top) extracts semantic hand states using a frozen HaMeR model and maps them to embodiment-specific finger joint angles via a learnable Retargeting Model. (2) The \textbf{SE(3) Estimation Branch} (bottom) utilizes a frozen 3D point tracker to recover the 6-DoF end-effector trajectory. The snowflake and flame icons indicate frozen pre-trained models and learnable modules, respectively.}
  \label{fig:dex_method_pipeline}
  \vspace{-3mm}
\end{figure}

TC-IDM is compatible with various dexterous hand tasks by only remapping the executor state, while keeping all upstream and downstream modules unchanged.  
In the following, we introduce a pipeline, as illustrated in Fig.~\ref{fig:dex_method_pipeline}, to estimate executor states and perform inverse dynamics for migrating human hand motions to different dexterous hands.

\begin{figure*}[h]
  \centering
  \includegraphics[width=\linewidth]{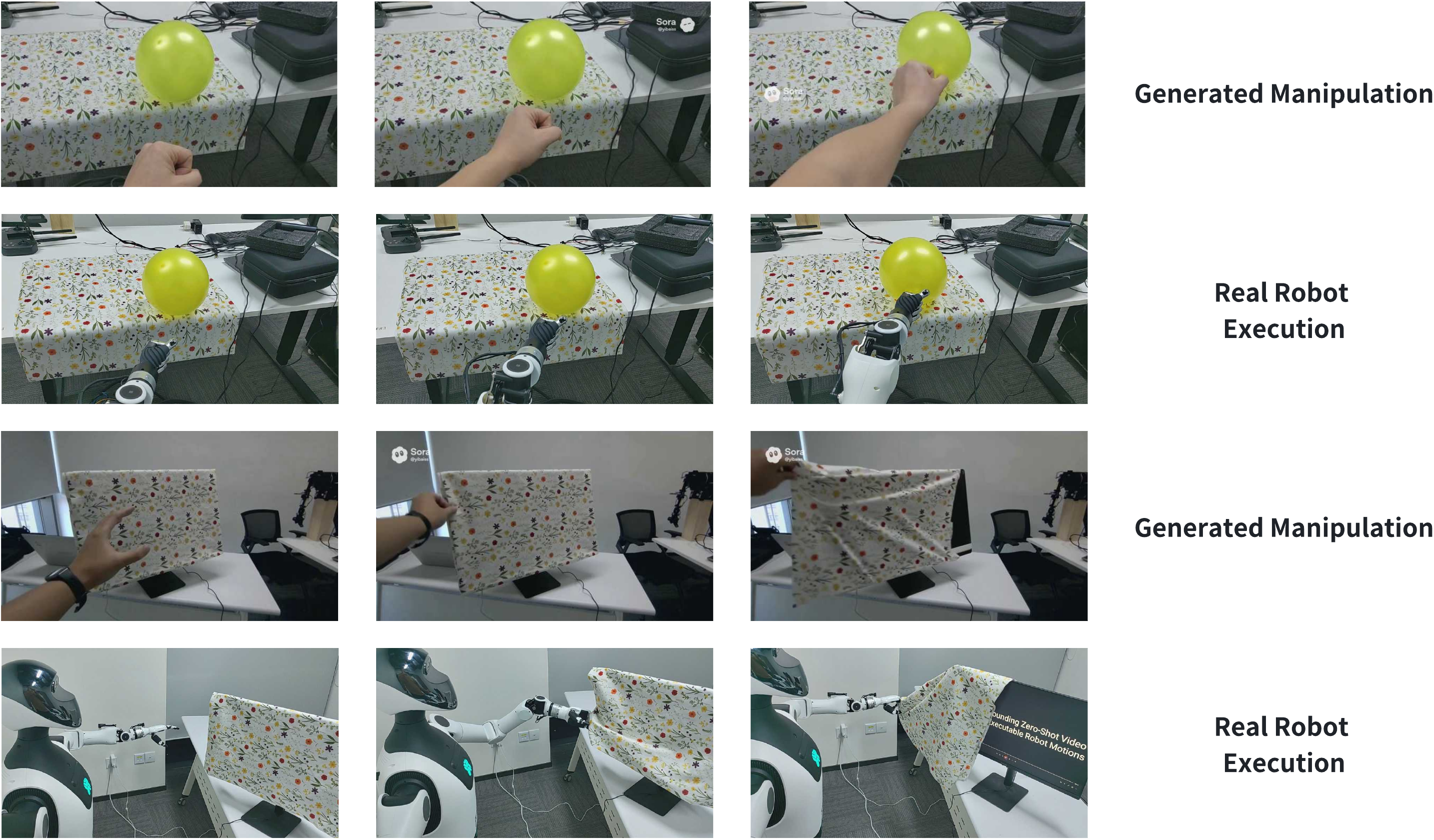} 
  \caption{\textbf{
  Qualitative results of TC-IDM in diverse generalization scenarios.} Our model successfully handles viewpoint shifts, long-horizon manipulation, deformable objects, and unseen tools. }
  \label{fig:dexterous_experiments}
  \vspace{-4mm}
\end{figure*}
\paragraph{Human Hand State Estimation.}
Given the generated RGB video
\begin{equation}
V_{\text{rgb-gen}} = \{ I_{\text{rgb-gen}}^t \}_{t=0}^T,
\end{equation}
We estimate human hand states using a frozen, pre-trained hand model \textbf{HaMeR}~\citep{pavlakos2024reconstructing}.  

At each time step, HaMeR predicts a 3D hand representation:
\begin{equation}
\mathcal{H}^t = \text{HaMeR}\!\left(I_{\text{rgb-gen}}^t\right),
\end{equation}
where $\mathcal{H}^t$ encodes hand pose and articulation relevant for grasping and manipulation.

\paragraph{Retargeting to Dexterous Hands.}
Instead of directly predicting robot joint angles, we introduce a retargeting network that maps human hand states to dexterous hand actions.  
For a given dexterous hand executor, the retargeting model predicts control commands as
\begin{equation}
A_{\text{dex}}^t = \mathcal{R}_{\text{hand}\rightarrow\text{dex}}\!\left(\mathcal{H}^t\right),
\end{equation}
where $A_{\text{dex}}^t$ denotes the joint command vector of the dexterous hand, and $\mathcal{R}_{\text{hand}\rightarrow\text{dex}}$ is a lightweight MLP trained for that embodiment.

This design decouples human hand understanding from robot kinematics. By only changing the retargeting head, the same human hand representation can drive different dexterous hands.

\paragraph{Experiments on Dexterous Hands}

We evaluate zero-shot transfer on two tasks: balloon hitting and cloth removal.
The two tasks use different dexterous hands: \textbf{BrainCo} for balloon hitting and \textbf{Inspire-Robots} for cloth removal. TC-IDM successfully transfers motions to both dexterous hands in both tasks.
These results show that TC-IDM learns an embodiment-agnostic motion representation and enables zero-shot retargeting to high-DoF dexterous hands without fine-tuning.

\section{Conclusion}

We present \textbf{TC-IDM}, a gripper-centric inverse dynamics model that bridges the gap between pixel-level visual planning and physically executable robot actions. By focusing on the imagined trajectory of the gripper, TC-IDM provides a robust and direct intermediate representation between generative world models and low-level control. Our decoupled "plan-and-translate" architecture effectively transforms visual–textual plans into executable manipulation trajectories, enhancing robustness to camera viewpoints, deformable object handling, and long-horizon generalization.

Through real-world experiments, we demonstrate that TC-IDM significantly improves execution performance, achieving 61.11\% average success across tasks and 38.46\% in zero-shot deformable object scenarios—substantially outperforming end-to-end VLA-style baselines. These results validate the effectiveness of gripper-centric modeling as a scalable and generalizable solution for grounded embodied control.

\clearpage
{\small
\bibliographystyle{ieeenat_fullname}
\bibliography{main}
}


\appendix
\clearpage
\setcounter{page}{1}

%

\section{Hardware Details}
\label{sec:Hardware}

\begin{figure}[h]
\centering
\includegraphics[width=\linewidth]{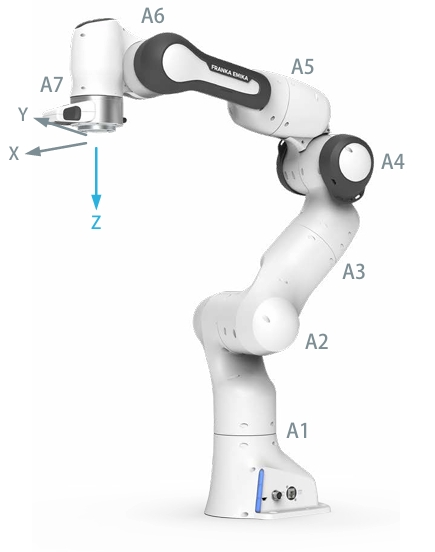}
\caption{Hardware features}
\label{fig:franka_case}
\vspace{-4mm}
\end{figure}

\textbf{Robotic Platform and End-Effector Configuration} The experimental evaluation was conducted using a single-arm Franka Emika Panda manipulatorFig.~\ref{fig:franka_case} integrated with a Robotiq 2F-85 adaptive gripper. The Franka Panda is a 7-DOF (degrees of freedom) collaborative robot designed for agile and sensitive manipulation. Unlike traditional position-controlled industrial robots, the Panda features high-resolution torque sensors at each joint, enabling direct torque control and compliant interaction behaviors. This architecture allows for precise impedance control, which is critical for unstructured environments where contact dynamics are variable.

The end-effector employed is the Robotiq 2F-85, a two-finger adaptive mechanical gripper. It features a stroke of $85$ mm and a payload capacity of $5$ kg. The gripper is mechanically coupled to the Panda’s flange and communicates via standard industrial protocols (Modbus RTU converted to USB/Ethernet). This configuration provides a versatile grasping solution capable of handling objects with varying geometries (encompassing/fingertip grips) and stiffness capabilities. The low-level control of the arm is managed via the Franka Control Interface (FCI), which exposes the robot state and allows for torque and position commands at a frequency of $1$ kHz, ensuring low-latency synchronization with the vision system.

\section{Dataset Details}
\label{sec:Dataset results}

To foster a robust assessment of manipulation capabilities, we constructed a large-scale training corpus derived from the Robomind dataset. As detailed in Table 1, the training set comprises $30,210$ demonstration trajectories spanning $25$ distinct tasks. These tasks are curated to encompass a wide manifold of manipulation semantics, ranging from elementary pick-and-place operations to complex, contact-rich fine-grained manipulations.To rigorously evaluate the agent's zero-shot generalization ability, we established a strictly held-out test set. The evaluation suite consists of $9$ unseen tasks that were explicitly excluded from the training distribution. To analyze performance across different complexity regimes, these test tasks are stratified into three difficulty levels (Easy, Medium, and Hard), with three tasks allocated to each level. This hierarchical structure allows us to disentangle the agent's capability in basic spatial reorientation from its proficiency in intricate object interaction.
\begin{table*}[h]
\centering
\caption{Overview of the dataset split and task difficulty distribution used in experiments.}
\label{tab:dataset_params}
\begin{tabular}{ll}
\toprule
\textbf{Parameter} & \textbf{Value} \\
\midrule
Training Dataset & Robomind Subset ($30,210$ trajectories, $25$ tasks) \\
Test Dataset & $9$ Unseen Tasks (Strictly held-out) \\
Test Difficulty Distribution & $3$ Easy, $3$ Medium, $3$ Hard \\
Data Exclusion & Test tasks are strictly excluded from training \\
\bottomrule
\end{tabular}
\end{table*}

\begin{figure*}[t]
    \centering
    \includegraphics[width=0.9\linewidth]{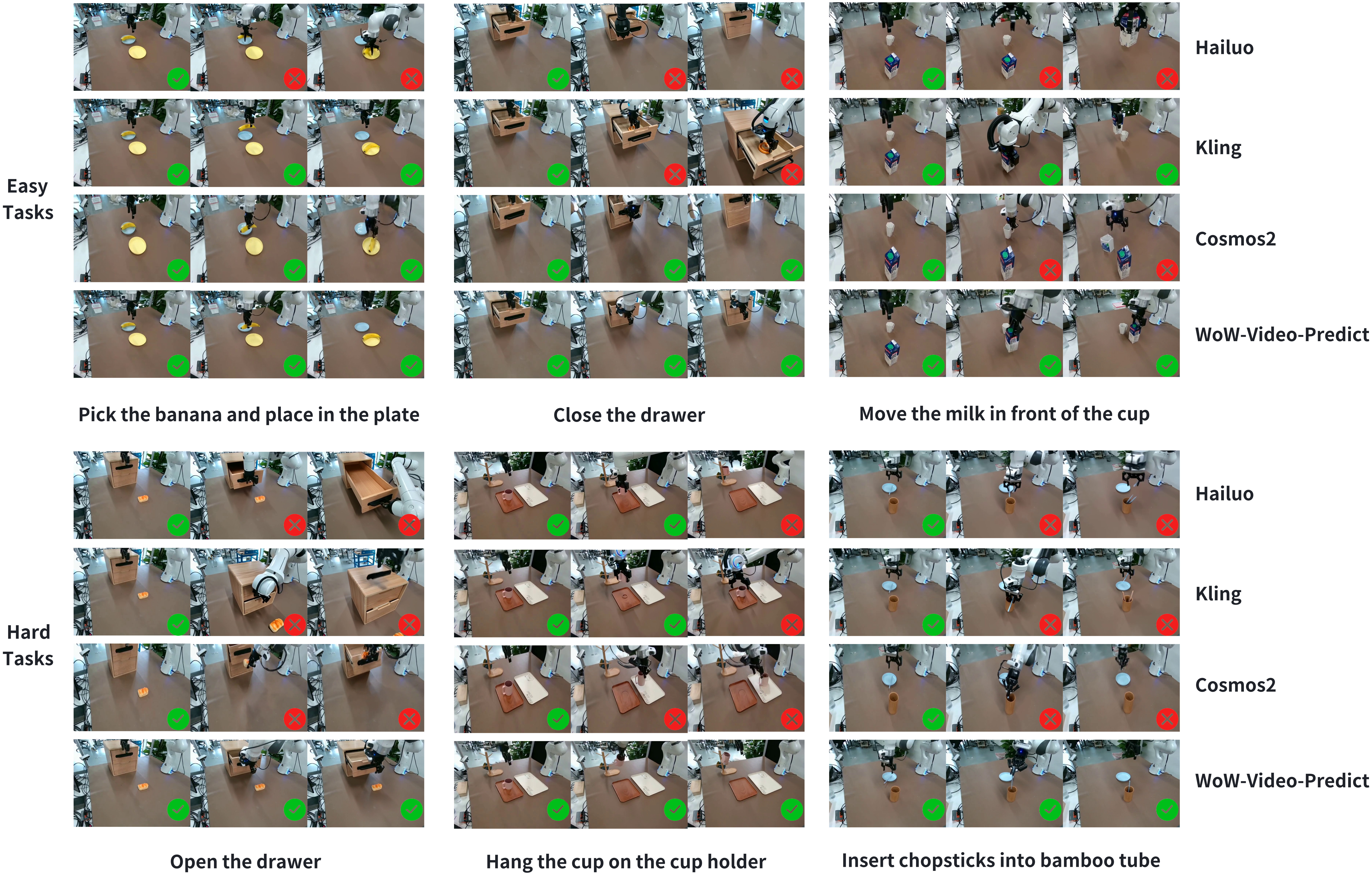}
    \caption{\textbf{Generated manipulation results for evaluating manipulation performance across different video models.}}
    \label{fig:vla_results_two_task}
\end{figure*}
\section{More Results}
\label{sec:more results}
Fig.~\ref{fig:vla_results_two_task} presents a comprehensive qualitative comparison of generated manipulation trajectories across four representative video generation models—Hailuo, Kling, Cosmos2, and WoW-Video-Predict—evaluated on both easy and hard manipulation tasks. Each row corresponds to a particular video model, while each column shows a distinct task scenario ranging from simple object relocation to complex tool-use or long-horizon interactions. The visual sequences illustrate how each model interprets the same initial scene and instruction, and how these interpretations unfold as predicted motion over time.

For easy tasks, such as picking a banana and placing it in a plate, closing a drawer, or moving milk in front of a cup, most models are able to produce semantically aligned motion plans with reasonable spatial consistency. Differences, however, can still be observed: some models demonstrate smoother gripper motion but weaker object-state continuity, while others maintain stronger physical plausibility yet occasionally deviate from the intended goal. Green markers indicate cases where the generated plan is visually plausible and goal-consistent, whereas red markers highlight clear failure modes such as misplaced trajectories, object hallucination, temporal jittering, or misalignment with the target instruction.

In contrast, hard tasks introduce more challenging dynamics and multi-step interactions, including opening a drawer, hanging a cup on a holder, and inserting chopsticks into a bamboo tube. These tasks require accurate 3D reasoning, fine manipulation skills, occlusion handling, and stable long-horizon motion prediction. The results reveal substantial variation among models: certain methods struggle to maintain consistent tool–object interaction, leading to drift or incomplete action sequences; others fail to preserve object geometry or relative spatial relations when handling tight clearances. Only a subset of models—most notably WoW-Video-Predict—produce coherent and physically plausible trajectories across the entire set of challenges.

Overall, this figure highlights the diverse generative behaviors of current video models when applied to robotic manipulation planning. The successes and failure cases observed here provide valuable insight into the reliability, temporal stability, and semantic grounding of visual plans produced by each model.

Additional examples, high-resolution renderings, and side-by-side comparisons with real-robot executions are provided in the accompanying video materials.

\end{document}